\documentclass{article}
\usepackage{ijcai26}
\usepackage[most]{tcolorbox}

\usepackage{times}
\usepackage{soul}
\usepackage{url}
\usepackage[hidelinks]{hyperref}
\usepackage[utf8]{inputenc}
\usepackage[small]{caption}
\usepackage{graphicx}
\usepackage{amsmath}
\usepackage{amsthm}
\usepackage{booktabs}
\usepackage{algorithm}
\usepackage{algorithmic}
\usepackage[switch]{lineno}
\usepackage{enumitem}
\usepackage{tabularx}
\usepackage{float}
\usepackage{placeins}
\usepackage{dblfloatfix}
\usepackage{multirow}
\usepackage{makecell}
\usepackage{threeparttable}
\usepackage{siunitx}
\usepackage{adjustbox}

\usepackage{xcolor}
\usepackage{comment}

\newcommand{\citep}[1]{\cite{#1}}
\newcommand{\citet}[1]{\citeauthor{#1}~\shortcite{#1}}

\newcommand{\StdGen}{Standard Instruction QA Data Generation}
\newcommand{\DiffGen}{Difficulty-Graded QA Data Generation}
\newcommand{\StdQA}{Standard Instruction QA Data}
\newcommand{\DiffQA}{Difficulty-Graded QA Data}

\newenvironment{promptbox}[1]{%
  \begin{quote}\small
  \noindent\textbf{#1}\par\smallskip\hrule\smallskip
}{%
  \smallskip\hrule
  \end{quote}
}

\usepackage{listings}

\usepackage{tikz}
\usetikzlibrary{shapes.geometric, arrows.meta, positioning, calc, fit, backgrounds, shapes.misc}

\usepackage{amssymb}

\usepackage{textcomp}

\definecolor{mainblue}{RGB}{218,235,250}
\definecolor{levelblue}{RGB}{190,220,245}
\definecolor{arrowcolor}{RGB}{70,70,70}
\definecolor{textblack}{RGB}{45,45,45}
\definecolor{textgray}{RGB}{100,100,100}
\definecolor{bordergray}{RGB}{175,175,175}

\definecolor{exBlue}{RGB}{232,241,255}
\definecolor{exBlue2}{RGB}{245,249,255}
\definecolor{exBorder}{RGB}{70,70,70}
\definecolor{exGray}{RGB}{248,248,248}

\definecolor{exBlue}{RGB}{232,241,255}
\definecolor{exBlue2}{RGB}{245,249,255}
\definecolor{exBorder}{RGB}{70,70,70}
\definecolor{tag1}{RGB}{76,114,176}   
\definecolor{tag2}{RGB}{85,168,104}   
\definecolor{tag3}{RGB}{196,78,82}    

\definecolor{exBorder}{RGB}{70,70,70}
\definecolor{exGray}{RGB}{250,250,250}

\definecolor{Lone}{RGB}{226,240,217}   
\definecolor{Ltwo}{RGB}{255,242,204}   
\definecolor{Lthree}{RGB}{242,220,219} 

\definecolor{tag1}{RGB}{56,118,29}
\definecolor{tag2}{RGB}{191,144,0}
\definecolor{tag3}{RGB}{192,0,0}

\definecolor{exFrame}{RGB}{120,120,120}
\definecolor{exBg}{RGB}{252,252,252}

\definecolor{L1c}{RGB}{34,139,34}   
\definecolor{L2c}{RGB}{200,140,0}   
\definecolor{L3c}{RGB}{200,40,40}   

\definecolor{hl}{RGB}{235,245,255}  

\tcbset{
  enhanced,
  boxrule=0.4pt,
  colframe=exFrame,
  colback=white,
  arc=2mm,
  left=6pt,
  right=6pt,
  top=5pt,
  bottom=5pt
}

\newtcolorbox{levelcard}[1]{%
  colback=white,
  colframe=black!25,
  boxrule=0.3pt,
  arc=2mm,
  borderline west={2.2pt}{0pt}{#1}
}

\lstdefinestyle{ijcaiwrap}{
  basicstyle=\ttfamily\footnotesize,
  columns=fullflexible,
  keepspaces=true,
  showstringspaces=false,
  breaklines=true,
  breakatwhitespace=false, 
  breakautoindent=true
}

\title{Domain-Adaptation through Synthetic Data:\\ Fine-Tuning Large Language Models for German Law}

\author{
Ali Hamza Bashir$^{1,2,3}$\and
Muhammad Rehan Khalid$^{2}$\and
Kostadin Cvejoski$^{5}$\and
Jana Birr$^{1}$\and \\
Jule Berghaus$^{4}$\and
Armin Berger$^{1,3,4}$\and
Sandra Halscheidt$^{1}$\and
Christian Temath$^{1}$\and\\
Rafet Sifa$^{1,3,4}$\and
David Berghaus$^{1,3}$\\
\affiliations
$^{1}$Fraunhofer IAIS\\
$^{2}$Georg-August-University Göttingen\\
$^{3}$Lamarr Institute\\
$^{4}$University of Bonn\\
$^{5}$JetBrains Research\\
}


\begin{document}


\maketitle

\begin{abstract}
Large language models (LLMs) often struggle in specialized domains such as legal reasoning due to limited expert knowledge, resulting in factually incorrect outputs or hallucinations. This paper presents an effective method for adapting advanced LLMs to German legal question answering through a novel synthetic data generation approach. In contrast to costly human-annotated resources or unreliable synthetic alternatives, our approach systematically produces high-quality, diverse, and legally accurate question-answer pairs directly from authoritative German statutes. Using rigorous automated filtering methods and parameter-efficient fine-tuning techniques, we demonstrate that LLMs adapted with our synthetic dataset significantly outperform their baseline counterparts on German legal question answering tasks. Our results highlight the feasibility of using carefully designed synthetic data as a robust alternative to manual annotation in high-stakes, knowledge-intensive domains.
\end{abstract}


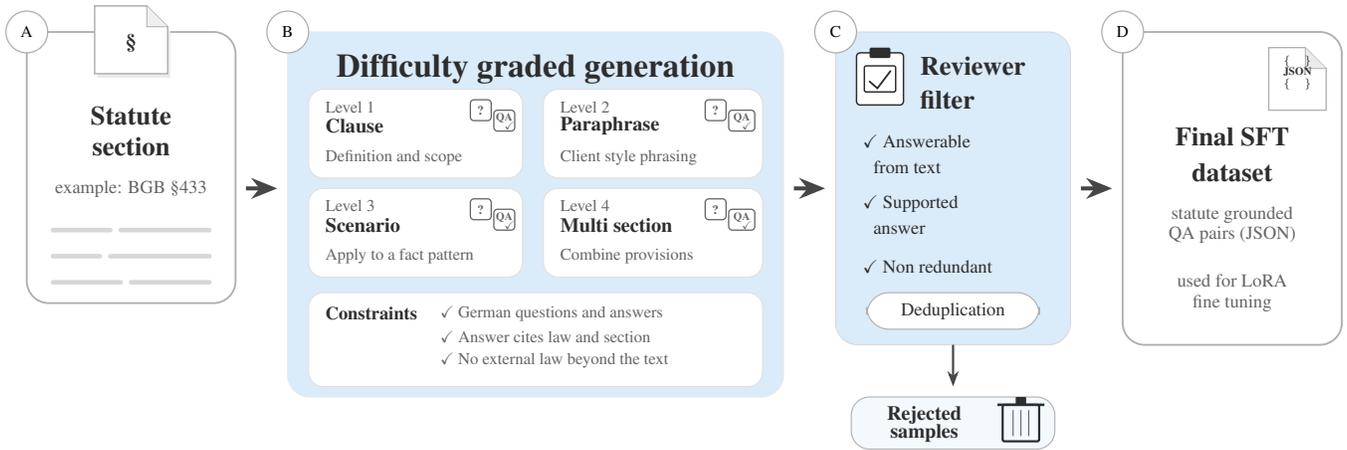
\begin{figure*}[t]
\centering
\resizebox{\textwidth}{!}{%
\begin{tikzpicture}[
    font=\sffamily,
    x=1cm, y=1cm,
    arrow/.style={
        ->,
        >={Stealth[length=5mm, width=4mm]},
        line width=2.2pt,
        arrowcolor
    },
    smallarrow/.style={
        ->,
        >={Stealth[length=3mm, width=2.2mm]},
        line width=1.2pt,
        arrowcolor
    }
]


\begin{scope}[shift={(0,0.3)}]
    \fill[white, rounded corners=12pt] (0,0.0) rectangle (4.0,5.2);
    \draw[bordergray, line width=1pt, rounded corners=12pt] (0,0.0) rectangle (4.0,5.2);
    
    \fill[black!8] (1.35,4.32) -- (1.35,5.62) -- (2.35,5.62) -- (2.75,5.22) -- (2.75,4.32) -- cycle;
    \fill[white] (1.3,4.4) -- (1.3,5.7) -- (2.3,5.7) -- (2.7,5.3) -- (2.7,4.4) -- cycle;
    \draw[bordergray, line width=0.8pt] (1.3,4.4) -- (1.3,5.7) -- (2.3,5.7) -- (2.7,5.3) -- (2.7,4.4) -- cycle;
    \fill[bordergray!20] (2.3,5.7) -- (2.3,5.3) -- (2.7,5.3) -- cycle;
    \draw[bordergray, line width=0.5pt] (2.3,5.7) -- (2.3,5.3) -- (2.7,5.3);
    \node[font=\Large\bfseries, text=textblack] at (2.0,5.0) {\textsection};
    
    \node[circle, draw=bordergray, fill=white, line width=0.7pt, minimum size=8mm, font=\small, inner sep=0pt] 
        at (0.0,5.2) {A};
    
    \node[font=\LARGE\bfseries, text=textblack] at (2.0,3.6) {Statute};
    \node[font=\LARGE\bfseries, text=textblack] at (2.0,3.0) {section};
    
    \node[font=\normalsize, text=textgray] at (2.0,2.2) {example: BGB \textsection433};
    
    \draw[bordergray!50, line width=2.5pt, line cap=round] (0.5,1.4) -- (1.6,1.4);
    \draw[bordergray!50, line width=2.5pt, line cap=round] (1.8,1.4) -- (3.5,1.4);
    \draw[bordergray!50, line width=2.5pt, line cap=round] (0.5,0.9) -- (1.4,0.9);
    \draw[bordergray!50, line width=2.5pt, line cap=round] (1.6,0.9) -- (3.5,0.9);
    \draw[bordergray!50, line width=2.5pt, line cap=round] (0.5,0.4) -- (1.8,0.4);
    \draw[bordergray!50, line width=2.5pt, line cap=round] (2.0,0.4) -- (3.5,0.4);
    
\end{scope}

\draw[arrow] (4.2,2.5) -- (4.8,2.5);


\begin{scope}[shift={(5.0,-0.5)}]
    \fill[mainblue, rounded corners=12pt] (0,-1.0) rectangle (9.5,6.0);
    \draw[bordergray!30, line width=0.5pt, rounded corners=12pt] (0,-1.0) rectangle (9.5,6.0);
    
    \node[font=\fontsize{18}{20}\selectfont\bfseries, text=textblack] at (4.75,5.3) {Difficulty graded generation};
    
    \node[circle, draw=bordergray, fill=white, line width=0.7pt, minimum size=8mm, font=\small, inner sep=0pt] 
        at (0.0,6.0) {B};
    
    \fill[white, rounded corners=8pt] (0.4,3.2) rectangle (4.5,4.9);
    \draw[bordergray!40, line width=0.5pt, rounded corners=8pt] (0.4,3.2) rectangle (4.5,4.9);
    \node[font=\small, text=textgray, anchor=north west] at (0.6,4.8) {Level 1};
    \node[font=\large\bfseries, text=textblack, anchor=west] at (0.6,4.2) {Clause};
    \node[font=\small, text=textgray, anchor=west] at (0.6,3.6) {Definition and scope};
    \draw[textgray, line width=0.7pt, rounded corners=2pt] (3.5,4.3) rectangle (3.9,4.7);
    \node[font=\scriptsize\bfseries, text=textgray] at (3.7,4.5) {?};
    \draw[textgray, line width=0.7pt, rounded corners=2pt] (3.95,4.1) rectangle (4.35,4.5);
    \node[font=\tiny\bfseries, text=textgray] at (4.15,4.35) {QA};
    \node[font=\tiny, text=textgray] at (4.25,4.2) {\checkmark};
    
    \fill[white, rounded corners=8pt] (4.9,3.2) rectangle (9.1,4.9);
    \draw[bordergray!40, line width=0.5pt, rounded corners=8pt] (4.9,3.2) rectangle (9.1,4.9);
    \node[font=\small, text=textgray, anchor=north west] at (5.1,4.8) {Level 2};
    \node[font=\large\bfseries, text=textblack, anchor=west] at (5.1,4.2) {Paraphrase};
    \node[font=\small, text=textgray, anchor=west] at (5.1,3.6) {Client style phrasing};
    \draw[textgray, line width=0.7pt, rounded corners=2pt] (8.0,4.3) rectangle (8.4,4.7);
    \node[font=\scriptsize\bfseries, text=textgray] at (8.2,4.5) {?};
    \draw[textgray, line width=0.7pt, rounded corners=2pt] (8.45,4.1) rectangle (8.95,4.5);
    \node[font=\tiny\bfseries, text=textgray] at (8.7,4.35) {QA};
    \node[font=\tiny, text=textgray] at (8.8,4.2) {\checkmark};
    
    \fill[white, rounded corners=8pt] (0.4,1.3) rectangle (4.5,3.0);
    \draw[bordergray!40, line width=0.5pt, rounded corners=8pt] (0.4,1.3) rectangle (4.5,3.0);
    \node[font=\small, text=textgray, anchor=north west] at (0.6,2.9) {Level 3};
    \node[font=\large\bfseries, text=textblack, anchor=west] at (0.6,2.3) {Scenario};
    \node[font=\small, text=textgray, anchor=west] at (0.6,1.7) {Apply to a fact pattern};
    \draw[textgray, line width=0.7pt, rounded corners=2pt] (3.5,2.4) rectangle (3.9,2.8);
    \node[font=\scriptsize\bfseries, text=textgray] at (3.7,2.6) {?};
    \draw[textgray, line width=0.7pt, rounded corners=2pt] (3.95,2.2) rectangle (4.35,2.6);
    \node[font=\tiny\bfseries, text=textgray] at (4.15,2.45) {QA};
    \node[font=\tiny, text=textgray] at (4.25,2.3) {\checkmark};
    
    \fill[white, rounded corners=8pt] (4.9,1.3) rectangle (9.1,3.0);
    \draw[bordergray!40, line width=0.5pt, rounded corners=8pt] (4.9,1.3) rectangle (9.1,3.0);
    \node[font=\small, text=textgray, anchor=north west] at (5.1,2.9) {Level 4};
    \node[font=\large\bfseries, text=textblack, anchor=west] at (5.1,2.3) {Multi section};
    \node[font=\small, text=textgray, anchor=west] at (5.1,1.7) {Combine provisions};
    \draw[textgray, line width=0.7pt, rounded corners=2pt] (8.0,2.4) rectangle (8.4,2.8);
    \node[font=\scriptsize\bfseries, text=textgray] at (8.2,2.6) {?};
    \draw[textgray, line width=0.7pt, rounded corners=2pt] (8.45,2.2) rectangle (8.95,2.6);
    \node[font=\tiny\bfseries, text=textgray] at (8.7,2.45) {QA};
    \node[font=\tiny, text=textgray] at (8.8,2.3) {\checkmark};
    
    \fill[white, rounded corners=8pt] (0.4,-0.8) rectangle (9.1,1.0);
    \draw[bordergray!40, line width=0.5pt, rounded corners=8pt] (0.4,-0.8) rectangle (9.1,1.0);
    \node[font=\normalsize\bfseries, text=textblack, anchor=west] at (0.6,0.6) {Constraints};
    \node[font=\small, text=textgray, anchor=west] at (2.8,0.6) {\checkmark\ German questions and answers};
    \node[font=\small, text=textgray, anchor=west] at (2.8,0.15) {\checkmark\ Answer cites law and section};
    \node[font=\small, text=textgray, anchor=west] at (2.8,-0.3) {\checkmark\ No external law beyond the text};
    
\end{scope}

\draw[arrow] (14.7,2.5) -- (15.3,2.5);


\begin{scope}[shift={(15.5,-0.5)}]
    \fill[mainblue, rounded corners=12pt] (0,0) rectangle (4.5,6.0);
    \draw[bordergray!50, line width=0.5pt, rounded corners=12pt] (0,0) rectangle (4.5,6.0);
    
    \fill[white, rounded corners=2pt] (0.4,4.6) rectangle (1.3,5.6);
    \draw[textblack, line width=0.8pt, rounded corners=2pt] (0.4,4.6) rectangle (1.3,5.6);
    \fill[textblack, rounded corners=1pt] (0.65,5.5) rectangle (1.05,5.7);
    \draw[textblack, line width=0.8pt, rounded corners=1pt] (0.55,4.85) rectangle (1.15,5.35);
    \draw[textblack, line width=1pt, line cap=round] (0.65,5.05) -- (0.8,4.92) -- (1.05,5.25);
    
    \node[font=\LARGE\bfseries, text=textblack, anchor=west] at (1.5,5.35) {Reviewer};
    \node[font=\LARGE\bfseries, text=textblack, anchor=west] at (1.5,4.7) {filter};
    
    \node[circle, draw=bordergray, fill=white, line width=0.7pt, minimum size=8mm, font=\small, inner sep=0pt] 
        at (0.0,6.0) {C};
    
    \node[font=\normalsize, text=textblack, anchor=west] at (0.4,3.9) {\checkmark\ Answerable};
    \node[font=\normalsize, text=textblack, anchor=west] at (0.6,3.4) {from text};
    \node[font=\normalsize, text=textblack, anchor=west] at (0.4,2.7) {\checkmark\ Supported};
    \node[font=\normalsize, text=textblack, anchor=west] at (0.6,2.2) {answer};
    \node[font=\normalsize, text=textblack, anchor=west] at (0.4,1.5) {\checkmark\ Non redundant};
    
    \fill[white, rounded corners=12pt] (0.6,0.3) rectangle (3.9,1.0);
    \draw[bordergray, line width=0.6pt, rounded corners=12pt] (0.6,0.3) rectangle (3.9,1.0);
    \node[font=\normalsize, text=textblack] at (2.25,0.65) {Deduplication};
    
\end{scope}

\draw[arrow] (20.2,2.5) -- (20.8,2.5);


\begin{scope}[shift={(21.0,-0.5)}]
    \fill[white, rounded corners=12pt] (0,0) rectangle (4.2,6.0);
    \draw[bordergray, line width=1pt, rounded corners=12pt] (0,0) rectangle (4.2,6.0);
    
    \fill[white] (2.8,4.5) -- (2.8,5.7) -- (3.5,5.7) -- (3.9,5.3) -- (3.9,4.5) -- cycle;
    \draw[bordergray, line width=0.7pt] (2.8,4.5) -- (2.8,5.7) -- (3.5,5.7) -- (3.9,5.3) -- (3.9,4.5) -- cycle;
    \fill[bordergray!25] (3.5,5.7) -- (3.5,5.3) -- (3.9,5.3) -- cycle;
    \draw[bordergray, line width=0.4pt] (3.5,5.7) -- (3.5,5.3) -- (3.9,5.3);
    \node[font=\tiny, text=textblack] at (3.15,5.45) {\{};
    \node[font=\tiny\bfseries, text=textblack] at (3.35,5.25) {JSON};
    \node[font=\tiny, text=textblack] at (3.55,5.45) {\}};
    \node[font=\tiny, text=textblack] at (3.15,5.0) {\{};
    \node[font=\tiny, text=textblack] at (3.55,5.0) {\}};
    
    \node[circle, draw=bordergray, fill=white, line width=0.7pt, minimum size=8mm, font=\small, inner sep=0pt] 
        at (0.0,6.0) {D};
    
    \node[font=\LARGE\bfseries, text=textblack] at (2.1,4.0) {Final SFT};
    \node[font=\LARGE\bfseries, text=textblack] at (2.1,3.3) {dataset};
    
    \node[font=\normalsize, text=textgray, align=center] at (2.1,2.3) {statute grounded\\QA pairs (JSON)};
    
    \node[font=\normalsize, text=textgray, align=center] at (2.1,1.0) {used for LoRA\\fine tuning};
    
\end{scope}


\begin{scope}[shift={(15.5,-0.5)}]
    \draw[smallarrow] (2.25,0.0) -- (2.25,-0.8);
    
    \fill[mainblue!30, rounded corners=10pt] (0.3,-2.0) rectangle (4.2,-1.0);
    \draw[bordergray, line width=0.6pt, rounded corners=10pt] (0.3,-2.0) rectangle (4.2,-1.0);
    \node[font=\large\bfseries, text=textblack] at (1.7,-1.35) {Rejected};
    \node[font=\large\bfseries, text=textblack] at (1.7,-1.7) {samples};
    
    \draw[textblack, line width=1.2pt, rounded corners=1pt] (3.2,-1.85) rectangle (3.9,-1.15);
    \draw[textblack, line width=1.4pt] (3.1,-1.15) -- (4.0,-1.15);
    \fill[textblack, rounded corners=0.5pt] (3.45,-1.15) rectangle (3.65,-1.0);
    \draw[textgray, line width=0.6pt] (3.4,-1.75) -- (3.4,-1.25);
    \draw[textgray, line width=0.6pt] (3.55,-1.75) -- (3.55,-1.25);
    \draw[textgray, line width=0.6pt] (3.7,-1.75) -- (3.7,-1.25);
\end{scope}

\end{tikzpicture}%
}
\caption{\textbf{\DiffQA{} pipeline.} A statute section serves as ground truth input. The generator produces complementary question styles at increasing difficulty (clause, paraphrase, scenario, multi-section) while enforcing statute-only constraints. A reviewer model filters unanswerable, unsupported, or redundant pairs and performs deduplication. The retained question–answer pairs constitute the supervised fine-tuning (SFT) dataset used for LoRA adaptation.}
\label{fig:pipeline}
\end{figure*}

\section{Introduction}

LLMs achieve strong performance on many general NLP tasks, yet reliability remains a key obstacle in high-stakes domains such as law. Legal question answering requires faithful grounding in authoritative sources and careful handling of legal references. Hallucinations or unsupported claims can therefore be harmful \citep{hu-etal-2025-factual-law,dahl-etal-2024-largelegalfictions}.

Building reliable legal QA systems is particularly challenging outside English. Multilingual legal benchmarks show substantial performance gaps across languages and tasks, reflecting an ecosystem in which resources, benchmarks, and model developments have been disproportionately English-centric \citep{niklaus-etal-2023-lextreme}. For German civil law, \citep{buttner-habernal-2024-gerlayqa} introduce GerLayQA, a valuable dataset grounded in the \emph{Bürgerliches Gesetzbuch} (BGB). However, this dataset has been crawled from a legal forum and the licensing is therefore debatable, which limits its practical use. As a result, there is a practical need for scalable, statute-grounded training data that can be used to adapt open LLMs to German law.

We address this need by synthesizing German legal instruction data directly from statutory text and therefore avoid any licensing issues. Using a curated corpus of German laws \citep{elra-2023-legalmc4}, we generate question-answer pairs that are explicitly grounded in specific provisions.

However, such an approach is challenging because we need to synthesize data that is diverse enough to capture real-world scenarios that have not been encountered during training and additionally not destroy the general purpose performance of the model. Moreover, we found that naive data generation approaches can even lead to diminishing performance on validation datasets that are within the distribution of the training data.

We explore two different data generation approaches: (i) a simple instruction-style generation setup and (ii) a difficulty-graded strategy that elicits questions ranging from direct lookups to scenario-based and multi-section reasoning, inspired by self-instruction style pipelines \citep{taori-etal-2023-alpaca}. To improve factual consistency, we add an LLM-based review stage that filters unanswerable, unsupported, or redundant QA pairs.

Using this statute-grounded data, we adapt two modern open-source LLMs to German legal QA. Namely, we fine-tune LLaMA~3.1 (8B)\footnote{\url{https://huggingface.co/meta-llama/Llama-3.1-8B-Instruct}} and Gemma~3 (12B)\footnote{\url{https://huggingface.co/google/gemma-3-12b-it}} using parameter-efficient LoRA \citep{hu-etal-2021-lora}. 

Due to the lack of evaluation datasets for instruction-tuned LLMs on German legal tasks, we created four different evaluation datasets. Two datasets are based on the German-civil-law (BGB). One of these is created in a multiple-choice format, avoiding any bias in the grading protocol. The second one contains open-ended questions which we grade using an LLM-as-a-judge approach \citep{zheng-etal-2023-llmjudge,liu-etal-2023-geval}. 

While we ensure that the evaluation dataset only contains questions that were not seen during training, we also want to investigate the performance of the models in real-world scenarios that are completely out-of-distribution. For this, we create two evaluation dataset variants (again a multiple-choice and an open-ended one) based on the German subset of LegalMC4 \citep{elra-2023-legalmc4}, which is a crawled collection of legal documents from the internet.

Remarkably, across all test sets, the difficulty-graded generation plus filtering yields consistent gains over both the base models and the simpler generation baseline while preserving general capabilities.

Our contributions are:
\begin{itemize}[leftmargin=*]
\item We propose a practical pipeline to synthesize German legal QA data for instruction-tuning. Our pipeline works directly from legislative texts, avoiding the reuse and privacy limitations common to web-crawled datasets.
\item We introduce a difficulty-graded generation strategy together with an LLM-based filtering stage that improves grounding and factual consistency of synthetic QA pairs.
\item We provide an extensive evaluation of two open LLMs adapted to German law, demonstrating substantial accuracy improvements on held-out German legal QA benchmarks with no noticeable degradation on general performance.
\end{itemize}

The fine-tuned models and synthetic datasets will be released upon acceptance of this work.

\section{Related Work}

\paragraph{Legal-domain language models and grounding.}
Legal NLP has long benefited from domain adaptation of general-purpose language models. Early work focused on continued pre-training of encoder models on legal corpora, e.g., Legal-BERT \citep{chalkidis-etal-2020-legalbert} and CaseLaw-BERT \citep{zheng-etal-2021-caselay}, improving performance on classification, retrieval, and judgment-related tasks within specific jurisdictions.
With the shift toward instruction-tuned generative LLMs, recent work has specialized models for legal assistance via continued pre-training or supervised instruction tuning. Examples for Chinese law include \citep{guo-etal-2023-chatlaw,song-2023-lawgpt} and examples for compliance checks include \citep{thiago}.
A complementary line of research targets reliability through \emph{grounding}, for instance via retrieval augmented generation (RAG) that conditions generation on external legal sources such as statutes or case repositories \citep{huang-etal-2023-lawyer-llama,chen-etal-2023-lawyergpt,lars}.
Despite these advances, factual errors and hallucinations remain central concerns in legal generation, motivating stronger adaptation and verification techniques \citep{dahl-etal-2024-largelegalfictions,hu-etal-2025-factual-law}.

\paragraph{German legal NLP resources and multilingual benchmarks.}
Compared to English and Chinese, publicly reusable resources for German legal QA remain limited.
GerLayQA \citep{buttner-habernal-2024-gerlayqa} constitutes an important step by providing German layperson questions with expert answers grounded in the BGB.
However, forum-derived datasets can face practical reuse constraints (e.g., licensing and privacy considerations), and may not always be suitable as training data for broad model adaptation.
More broadly, multilingual legal benchmarks such as LEXTREME \citep{niklaus-etal-2023-lextreme} highlight persistent performance gaps across languages (including German), but focus primarily on tasks such as classification and extraction rather than open-ended, statute-grounded answer generation.
These limitations motivate approaches that create scalable supervision specifically aligned with German statute-based QA.

\paragraph{Training on synthetic data.}
Training on synthetic data is a common concept in machine learning and has been applied in a variety of contexts such as tabular data \citep{hollmann2023tabpfn,hollmann2025tabpfn}, robotics \citep{8202133} and time-series \citep{fim_mjp,fim_pp,fim_sde}.

In the context of LLMs, a common strategy is \emph{conditional synthetic data generation}. This means that a raw corpus is used as input for a generative model which then synthesizes the data.  Self-Instruct \citep{wang-etal-2023-self-instruct} formalizes this LLM-based generation of instruction--response pairs combined with filtering to remove invalid or redundant samples.
To systematically increase task complexity, Evol-Instruct-style methods (e.g., as used in WizardLM ) rewrite prompts into progressively more challenging variants, exposing models to a broader difficulty spectrum \citep{xu-etal-2023-wizardlm}.
Closely related to our setting, recent work shows that transforming domain corpora into reading-comprehension style QA can be an effective alternative to continued pre-training on raw text, while preserving instruction-following behavior \citep{cheng-etal-2024-readingcomprehension}.
Our pipeline follows this direction by converting statutory passages into QA pairs, explicitly controlling difficulty (from direct lookups to multi-provision reasoning) and applying reviewer-style filtering to improve statute-only grounding and factual consistency. 

\paragraph{Benchmarks and evaluation protocols.}
General legal NLP benchmarks such as LexGLUE \citep{chalkidis-etal-2022-lexglue} and LegalBench \citep{zheng-etal-2023-legalbench} provide broad coverage of legal tasks, but are English-centric and classification-oriented, so they do not directly measure German statute-grounded answer generation.
Since our target setting is German legal QA grounded in statutory text, we evaluate on held-out test sets derived from German statutes and LegalMC4 \citep{elra-2023-legalmc4}, matching the intended deployment scenario.
For open-ended QA, we adopt LLM-as-a-judge scoring as a scalable evaluation protocol \citep{zheng-etal-2023-llmjudge,liu-etal-2023-geval} and complement it with exact accuracy for multiple-choice questions to ensure robustness across evaluation formats.

\section{Data Preparation}
\label{sec:data-prep}

Our main contribution is the construction of statute-grounded supervised fine-tuning (SFT) data for German legal question answering.
The design goals are:
(i) \emph{grounding} in authoritative legal text,
(ii) \emph{coverage} of question types that resemble real user queries,
(iii) \emph{scalability} without manual annotation, and
(iv) \emph{contamination control} through passage-level splitting.
We operationalize these goals with two synthetic data pipelines.
\textbf{\StdGen{}} is a single-prompt baseline that converts a passage into instruction-style QA in one step.
\textbf{\DiffGen{}} is a difficulty-graded statute QA pipeline that generates multiple controlled question types per provision and applies stricter automated filtering.

\subsection{Source Texts and Normalization}
\label{sec:source-texts}

We build training supervision from German legal text with explicit provenance and identifiers. We use the statutory texts from the most common german laws, namely the \emph{Bürgerliches Gesetzbuch} (BGB) which is the German Civil Code, the \emph{Strafgesetzbuch} (StGB) which is the Criminal Code, the \emph{Sozialgesetzbuch} (SGB) which is the Social Code, and the \emph{Grundgesetz} (GG) which is the Basic Law.
For statutory material, we normalize each provision into a canonical record:
\begin{itemize}[leftmargin=*, itemsep=0.2em]
  \item \textbf{law identifier} (e.g., BGB, StGB, SGB, GG),
  \item \textbf{section identifier} (e.g., \S~433),
  \item \textbf{section text} (German).
\end{itemize}
This representation ensures that generated answers can explicitly reference the correct authority (law + section).
For evaluation set construction (Section~\ref{sec:experiments}), we additionally use the German portion of LegalMC4 \citep{elra-2023-legalmc4}, which contains heterogeneous legal web documents.
Across sources, we preserve the original passage boundaries and identifiers so that every QA pair can be audited against a fixed input span.

\subsection{\StdGen{}}
\label{sec:simple-synth}

\StdGen{} is a straightforward "passage-to-QA" baseline.

Given a single input passage, we prompt GPT-4 to generate a small set of German question--answer pairs and to return them in a strict JSON schema (Appendix~\ref{appendix:synthetic-gen-prompt}).
This setup follows a widely used synthetic supervision paradigm where an LLM converts raw text into task examples that can be used for supervised fine-tuning.
In particular, it is closely related to corpus-to-QA conversion approaches that transform domain corpora into reading-comprehension style supervision rather than continuing training on raw text \citep{cheng-etal-2024-readingcomprehension}.
It is also consistent with the general generate-and-filter recipe for building synthetic instruction data at scale \citep{wang-etal-2023-self-instruct}.

Concretely, the prompt enforces the following constraints:
\begin{itemize}[leftmargin=*, itemsep=0.2em]
  \item \textbf{Input:} one legal passage with its provenance (law identifier and section, when available).
  \item \textbf{Output:} up to $k$ QA pairs in German, formatted as JSON.
  \item \textbf{Style:} questions reflect plausible user queries, answers use a lawyer-style tone.
\end{itemize}

\StdGen{} is straightforward to implement, but it is unclear if such a naive and open-ended prompt yields data with sufficient diversification.
This motivates \DiffGen{}, which explicitly controls question type and difficulty and applies stricter review-based filtering (Section~\ref{sec:difficulty-synth}).

\subsection{\DiffGen{}}
\label{sec:difficulty-synth}

\DiffGen{} is designed to produce statute-grounded supervision that is both diverse and controllable.

Rather than generating one undifferentiated set of questions per provision, we explicitly target multiple question families that reflect how legal queries arise in practice.
The design is motivated by two observations.
First, synthetic supervision improves when the data covers a range of input--output patterns and difficulty levels \citep{bengio-etal-2009-curriculum}.
Second, instruction generation pipelines benefit from structured transformations that increase complexity while maintaining validity \citep{xu-etal-2023-wizardlm}.
Operationally, we generate questions at four difficulty levels and then apply a separate reviewer prompt during filtering (Appendix~\ref{appendix:prompts}).

Across all levels, we enforce shared invariants:
\begin{itemize}[leftmargin=*, itemsep=0.2em]
  \item \textbf{Statute-only grounding:} answers must be fully supported by the provided text and must not rely on external law or assumptions.
  \item \textbf{Explicit authority:} answers must cite the law identifier and the section number.
  \item \textbf{Controlled output:} the generator must return a strict JSON structure to enable automated parsing.
  \item \textbf{Redundancy control:} we cap the number of pairs per level to limit near-duplicate paraphrases.
\end{itemize}

\paragraph{Level 1 (clause-centric).}
Level~1 focuses on high-precision comprehension.
Questions address what the provision states, when it applies, and which conditions or exceptions are encoded in the text.
This level prioritizes faithfulness and encourages answers that correctly map to the cited authority.

\paragraph{Level 2 (client-style paraphrase).}
Level~2 aims to match how users typically ask legal questions.
By prohibiting statute identifiers in the question, we force the model to learn the mapping from natural language to the correct cited provision in the answer.
This reduces the risk that fine-tuning degenerates into surface-level pattern matching on section numbers.

\paragraph{Level 3 (scenario application).}
Level~3 introduces short, realistic fact patterns.
The question must be phrased independently of the statutory wording, which encourages application rather than restatement.
The answer must still be statute-only, and it must justify the conclusion by referencing the provision.

\paragraph{Level 4 (multi-provision reasoning).}
Level~4 targets composition across sections.
We provide two or more sections as input and require scenarios where multiple sections are substantively necessary.
To preserve grounding, the generator is instructed to output no pairs when a valid multi-section scenario cannot be constructed from the provided text alone.

\paragraph{Illustrative example.}
Figure~\ref{fig:approach1v2_example} illustrates the qualitative difference between \StdGen{} and \DiffGen{}. \StdGen{} produces single-level, recall-style supervision, whereas \DiffGen{} yields complementary, difficulty-graded supervision (L1--L3) from the same provision and enables multi-section reasoning at Level~4 when multiple inputs are provided, while maintaining statute-only grounding and explicit citation.

\begin{figure*}[t]
\centering
\setlength{\fboxsep}{7pt}
\setlength{\fboxrule}{0.6pt}

\fcolorbox{black!35}{white}{%
\begin{minipage}{0.97\textwidth}
\begingroup
\footnotesize
\setlength{\parindent}{0pt}
\setlength{\parskip}{0.35em}

\begingroup
\setlength{\fboxsep}{5pt}
\colorbox{black!10}{%
  \parbox{\dimexpr\linewidth-2\fboxsep\relax}{%
    \textbf{Input statute (English translation).}
    \textsc{BGB}~\S~857 (\textit{Inheritability of Possession}):
    \emph{Possession passes to the heir. English shown for readability; training data is generated in German.}
  }%
}
\endgroup

\vspace{0.9em}

\noindent
\begin{minipage}[t]{0.30\linewidth}
\vspace{0pt}
\raggedright
\textbf{\StdGen{}: Single-prompt generation (single level).}\par
{\footnotesize\emph{Typically produces short, statute-identification / recall-style supervision.}}\par
\vspace{0.6em}

\textbf{Q.} Which law and which provision governs the inheritability of possession?\par
\textbf{A.} The inheritability of possession is governed by \textsc{BGB}~\S~857.\par
\end{minipage}
\hfill
\begin{minipage}[t]{0.67\linewidth}
\vspace{0pt}
\raggedright
\textbf{\DiffGen{}: Difficulty-graded generation (L1--L4).}\par
{\footnotesize\emph{The same statute input yields progressively harder question types; Level~4 uses multiple sections as input.}}\par
\vspace{0.6em}

\begin{minipage}[t]{0.48\linewidth}
\vspace{0pt}
\raggedright
\textbf{L1 (clause comprehension).}\par
\textbf{Q.} What happens to the possession of a deceased person after their death?\par
\textbf{A.} According to \textsc{BGB}~\S~857, possession passes to the heir.\par
\end{minipage}
\hfill
\begin{minipage}[t]{0.48\linewidth}
\vspace{0pt}
\raggedright
\textbf{L2 (client-style paraphrase).}\par
\textbf{Q.} I am the sole heir of my deceased brother. Do I have to wait for court proceedings before I have access to his belongings?\par
\textbf{A.} According to \textsc{BGB}~\S~857, possession passes to the heir upon inheritance. You do not need to wait for court proceedings; as the heir, you immediately have legal possession of your brother's belongings.\par
\end{minipage}

\vspace{0.9em}

\textbf{L3 (scenario application; single-section input).}\par
\textbf{Q.} A tenant dies. The apartment remains locked, and the sole heir can only travel in a few days. Meanwhile, a close acquaintance removes several valuables and argues that the heir is not yet the ``possessor'' because he has not physically taken them. Is the heir already the legal possessor of the items in the apartment?\par
\textbf{A.} Yes. Under \textsc{BGB}~\S~857, possession passes to the heir. Thus, the heir becomes the possessor upon the inheritance event, even if he has not yet physically taken the items.\par

\vspace{0.8em}

{\footnotesize\emph{Example Level~4 input sections:} \textsc{BGB}~\S~857, \S~1362, \S~1384.}\par
\textbf{L4 (cross-section integration; multi-section input).}\par
\textbf{Q.} A married couple has been separated for months, and a divorce petition has already been filed. A creditor enforces against one spouse by seizing movable property located in the other spouse's apartment. Shortly thereafter, the debtor spouse dies; the heir demands immediate possession of the estate items. Which provisions govern (i) transfer of possession to the heir, (ii) the ownership presumption for seized movables, and (iii) the relevant date for calculating accrued gains?\par
\textbf{A.} (i) \textsc{BGB}~\S~857: possession passes to the heir upon inheritance. (ii) \textsc{BGB}~\S~1362: presumption in favor of the creditor that movables in spouses' possession belong to the debtor, with an exception when spouses live separately and the items are with the non-debtor spouse. (iii) \textsc{BGB}~\S~1384: the relevant date for calculating accrued gains is the pendency of the divorce petition.\par

\end{minipage}

\endgroup
\end{minipage}%
}

\caption{\textbf{\StdQA{} vs.\ \DiffQA{} example (English shown for readability).} \StdGen{} produces single-level, largely recall-style supervision from a statute section. \DiffGen{} generates difficulty-graded question types (L1--L3) from the same input provision and supports cross-section integration at Level~4 when multiple provisions are provided as input.}

\label{fig:approach1v2_example}
\end{figure*}

\subsection{LLM-Based Quality Filtering and Redundancy Control}
\label{sec:quality-filter}

\DiffGen{} follows a generate-then-filter recipe that is standard in synthetic SFT construction \citep{wang-etal-2023-self-instruct}.
Importantly, \DiffGen{} combines two complementary mechanisms:
(i) \emph{difficulty-graded generation} that systematically increases diversity by eliciting qualitatively different question families (Levels~1--4; Figure~\ref{fig:approach1v2_example}),
and (ii) a \emph{statute-only reviewer} that removes invalid or low-quality instances.
The reviewer is strictly \emph{subtractive}: it can only discard candidates, not generate new question types.
Thus, the structured multi-level prompting drives diversity, while the reviewer primarily improves fidelity and reduces noise.

We do not apply an analogous reviewer stage to \StdGen{}, as it is intended as a minimal one-step passage-to-QA baseline.
Adding a second model-based filtering component to \StdGen{} would introduce an additional design axis and blur the baseline comparison; we therefore keep \StdGen{} as generated.

For \DiffGen{}, we apply a dedicated reviewer prompt (Appendix~\ref{appendix:prompts}) that has access \emph{only} to the statute text used for generation.

For each candidate QA pair, the reviewer verifies three criteria:
\begin{itemize}[leftmargin=*, itemsep=0.2em]
  \item \textbf{Answerability:} the question is specific and can be answered from the provided text,
  \item \textbf{Support:} the answer is fully supported and introduces no external law or assumptions,
  \item \textbf{Non-redundancy:} the pair is not duplicative of other pairs from the same passage.
\end{itemize}
Pairs that fail any criterion are discarded.

As shown in Table~\ref{tab:qa_distribution}, \DiffGen{} generates 31{,}777 candidates before review and retains 23{,}905 pairs after review.
Rejections are concentrated in the lower difficulty levels (Levels~1--2), where redundant paraphrases are most common, whereas Level~4 is unaffected (441/441 retained).
This indicates that the reviewer primarily removes noise and redundancy rather than suppressing the rare cross-section samples.

\subsection{Train-Test Split and Contamination Control}
\label{sec:splits}

To prevent memorization of identical source passages, we split at the \emph{passage level}.
For statutes, splitting is performed at the section identifier (\S), ensuring that all QA pairs generated from a given section belong to exactly one split.
For LegalMC4-derived material, we split by held-out source passages so that evaluation passages do not appear in any training set.

Our evaluation comprises four held-out test sets derived from the BGB and LegalMC4:
LegalMC4 QA ($n{=}732$), BGB QA ($n{=}715$), LegalMC4 MCQ ($n{=}1315$), and BGB MCQ ($n{=}1845$).
\DiffQA{} data is used exclusively for model adaptation and is never used to construct evaluation sets.

\section{Experiments and Results}
\label{sec:experiments}

We evaluate domain specialization and generalization for two instruction-tuned models, LLaMA~3.1 (8B) and Gemma~3 (12B),
comparing fine-tuned variants against their base counterparts.

\subsection{LoRA Fine-tuning Setup}
\label{sec:lora-setup}

For each base model, we perform supervised LoRA fine-tuning on the synthetic training sets produced by \textbf{\StdGen{}} and \textbf{\DiffGen{}} (Section~\ref{sec:data-prep}). To isolate the effect of the training data generation approach, we keep the LoRA configuration and optimization hyperparameters fixed across both approaches and across both model families. Table~\ref{tab:lora_hparams} summarizes the shared hyperparameters used in all fine-tuning runs. For Gemma, we disable the vision layers.

\newsavebox{\loraHparamsBox}
\begin{table}[t]
\centering
\sbox{\loraHparamsBox}{%
\small
\setlength{\tabcolsep}{5pt}
\begin{tabular}{@{}ll@{}}
\toprule
Parameter & Value \\
\midrule
Max sequence length & 2048 \\
Precision & bf16 \\
LoRA rank ($r$) & 16 \\
LoRA alpha ($\alpha$) & 32 \\
LoRA dropout & 0.05 \\
Adapter placement$^{*}$ & Attention + MLP modules \\
Random state & 3407 \\
Epochs & 7 \\
Learning rate & $1 \times 10^{-5}$ \\
Optimizer & \texttt{AdamW\_8bit} \\
Weight decay & 0.01 \\
LR scheduler & linear \\
Warmup ratio & 0.1 \\
\bottomrule
\end{tabular}%
}
\begin{minipage}{\wd\loraHparamsBox}
\usebox{\loraHparamsBox}
\caption{LoRA fine-tuning hyperparameters in Gemma-3-12B-IT and Llama-3.1-Instruct.}
\label{tab:lora_hparams}
\vspace{2pt}
{\scriptsize $^{*}$For Gemma: vision layers are disabled.}
\end{minipage}
\end{table}

\subsection{Compute and Training Time}
\label{sec:compute}

All LoRA fine-tuning runs were executed on a single \textsc{NVIDIA A100-SXM4 GPU} with 80\,GB memory.
Fine-tuning one model on one synthetic dataset took approximately 9--10 hours of wall-clock time, including evaluations on the validation set and checkpointing.

\subsection{Evaluation Tasks}
\label{sec:eval}

We evaluate on four held-out German legal test sets derived from the BGB and the German portion of LegalMC4 \citep{elra-2023-legalmc4}.
Two sets contain open-ended questions and are scored for factual correctness using an LLM judge (GPT-4.1),
following standard LLM-based evaluation protocols \citep{zheng-etal-2023-llmjudge,liu-etal-2023-geval} which yielded in high correlations with human experts in similar scenarios \citep{lars}.
Two sets contain multiple-choice questions and are evaluated by exact accuracy.

Critically, the LegalMC4-based evaluation sets are constructed from real-world legal documents crawled from the internet, including court decisions, legal opinions, administrative documents, and other heterogeneous legal texts. These documents naturally contain substantial noisy context-irrelevant passages, formatting artifacts, tangential discussions, and information peripheral to the core legal question. This design choice is deliberately aligned with practical deployment scenarios: in production systems, legal LLMs are predominantly deployed within RAG architectures, where a retrieval component first identifies potentially relevant documents, and the generator model must then locate the pertinent information within these retrieved passages \citep{huang-etal-2023-lawyer-llama,chen-etal-2023-lawyergpt}.

By evaluating on LegalMC4 data with its inherent noise, we indirectly assess the model's ability to function as the generator component in a RAG system. A fine-tuning approach should not only improve the model's legal knowledge but also its robustness to distractor information and its capacity to extract relevant facts from imperfect retrievals---capabilities that are essential for real-world legal assistance applications. The evaluation therefore tests whether adapted models can maintain accurate reasoning even when presented with the noisy, heterogeneous document contexts typical of RAG-based legal QA systems. Figure~\ref{fig:legalmc4_rag_example} provides an illustrative context-based instance in this setting: the model is given a retrieved legal passage (which may include irrelevant or noisy surrounding text) and must answer a user question by extracting and grounding the response in the provided context. This mirrors the generator role in RAG pipelines, where retrieval provides candidate evidence and the model must identify the relevant facts and produce a faithful answer.

In contrast, the BGB-based evaluation sets provide cleaner, statute-focused assessments that isolate legal reasoning capabilities on well-structured legislative text.

To summarize, our evaluation sets are: \textbf{Open-ended QA (LLM-judged correctness)}: \textbf{LegalMC4 QA} ($n{=}732$), open-ended questions derived from LegalMC4 passages, requiring extraction and reasoning over information embedded in noisy real-world legal documents; and \textbf{BGB QA} ($n{=}715$), open-ended questions grounded in held-out BGB sections. \textbf{Multiple-choice QA (accuracy)}: \textbf{LegalMC4 MCQ} ($n{=}1315$), multiple-choice questions derived from LegalMC4 material, testing document comprehension in noisy real-world contexts; and \textbf{BGB MCQ} ($n{=}1845$), multiple-choice questions targeting the BGB.

All test sets are constructed from held-out source passages that do not appear in any training set (Section~\ref{sec:splits}).

\begin{table*}[t]
\centering
\small
\setlength{\tabcolsep}{7pt}
\begin{threeparttable}
\caption{Accuracy on German legal QA benchmarks. Open QA columns report LLM-judged factual correctness (\%). MCQ columns report exact accuracy (\%). }
\label{tab:domain-results}
\begin{tabular}{
   l
   l
   S[table-format=2.1]
   S[table-format=2.1]
   S[table-format=2.1]
   S[table-format=2.1]
}
\toprule
\multirow{2}{*}{\textbf{Model}} & \multirow{2}{*}{\textbf{Train Data}} &
\multicolumn{2}{c}{\textbf{Open QA (Judge)}} &
\multicolumn{2}{c}{\textbf{MCQ (Accuracy)}} \\
\cmidrule(lr){3-4} \cmidrule(lr){5-6}
& & {\makecell{\textbf{LegalMC4}\\\textbf{QA}}}
  & {\makecell{\textbf{BGB}\\\textbf{QA}}}
  & {\makecell{\textbf{LegalMC4}\\\textbf{MCQ}}}
  & {\makecell{\textbf{BGB}\\\textbf{MCQ}}} \\
\midrule

\multirow{3}{*}{\makecell[l]{LLaMA~3.1\\(8B)}}
& --        & 43.0 & 39.2 & 60.9 & 64.0 \\
& Standard Instruction QA Data  & \multicolumn{1}{c}{35.4~{\scriptsize\textcolor{red}{$\downarrow$7.6}}}
              & \multicolumn{1}{c}{37.6~{\scriptsize\textcolor{red}{$\downarrow$1.6}}}
              & \multicolumn{1}{c}{59.9~{\scriptsize\textcolor{red}{$\downarrow$1.0}}}
              & \multicolumn{1}{c}{64.2~{\scriptsize\textcolor{green!60!black}{$\uparrow$0.2}}} \\
& Difficulty-Graded QA Data  & \multicolumn{1}{c}{55.4~{\scriptsize\textcolor{green!60!black}{$\uparrow$12.4}}}
              & \multicolumn{1}{c}{59.2~{\scriptsize\textcolor{green!60!black}{$\uparrow$20.0}}}
              & \multicolumn{1}{c}{66.1~{\scriptsize\textcolor{green!60!black}{$\uparrow$5.2}}}
              & \multicolumn{1}{c}{68.2~{\scriptsize\textcolor{green!60!black}{$\uparrow$4.2}}} \\
\midrule

\multirow{3}{*}{\makecell[l]{Gemma~3\\(12B)}}
& --        & 38.7 & 36.8 & 66.8 & 67.0 \\
& Standard Instruction QA Data  & \multicolumn{1}{c}{41.8~{\scriptsize\textcolor{green!60!black}{$\uparrow$3.1}}}
              & \multicolumn{1}{c}{46.6~{\scriptsize\textcolor{green!60!black}{$\uparrow$9.8}}}
              & \multicolumn{1}{c}{66.9~{\scriptsize\textcolor{green!60!black}{$\uparrow$0.1}}}
              & \multicolumn{1}{c}{67.1~{\scriptsize\textcolor{green!60!black}{$\uparrow$0.1}}} \\
& Difficulty-Graded QA Data  & \multicolumn{1}{c}{54.5~{\scriptsize\textcolor{green!60!black}{$\uparrow$15.8}}}
              & \multicolumn{1}{c}{76.4~{\scriptsize\textcolor{green!60!black}{$\uparrow$39.6}}}
              & \multicolumn{1}{c}{71.2~{\scriptsize\textcolor{green!60!black}{$\uparrow$4.4}}}
              & \multicolumn{1}{c}{75.1~{\scriptsize\textcolor{green!60!black}{$\uparrow$8.1}}} \\
\midrule

GPT-5-mini (min. reasoning) & -- & 70.1 & 62.7 & 70.9 & 75.8 \\
GPT-5 (min. reasoning)      & -- & 77.2 & 74.7 & 72.0 & 78.8 \\
\bottomrule
\end{tabular}
\begin{tablenotes}
\item Arrows indicate performance change compared to the base model: {\scriptsize\textcolor{green!60!black}{$\uparrow$}} denotes improvement, {\scriptsize\textcolor{red}{$\downarrow$}} denotes decrease, {\scriptsize\textcolor{black}{$\rightarrow$}} denotes no change. Values show the absolute difference in performance metrics.
\end{tablenotes}
\end{threeparttable}
\end{table*}

\subsection{Domain-Specific Performance}
\label{sec:domain-results}

Table~\ref{tab:domain-results} reports performance on the four German legal test sets. The results reveal a clear hierarchy in performance.

\paragraph{\DiffGen{} consistently delivers substantial improvements.}
The difficulty-graded generation strategy with LLM-based filtering yields the strongest gains across all benchmarks and both model families. For Gemma~3 (12B), the improvements are particularly striking: open-ended correctness increases by approximately 40 percentage points on BGB QA (+39.6) and about 16 points on LegalMC4 QA (+15.8). Multiple-choice accuracy improves by about 4--8 points (+4.4 on LegalMC4 MCQ and +8.1 on BGB MCQ). LLaMA~3.1 (8B) shows similar patterns with consistent gains across all test sets, with the largest improvement on BGB QA (+20.0 points) and strong gains on LegalMC4 QA (+12.4 points).

\paragraph{LegalMC4 as a proxy for retrieval-augmented QA with context.}
Because the LegalMC4-based benchmarks provide noisy, heterogeneous document context (Section~\ref{sec:eval}), the gains on LegalMC4 QA/MCQ indicate that Difficulty-Graded QA fine-tuning also improves performance in retrieval-augmented deployments where the model must answer using retrieved passages \citep{huang-etal-2023-lawyer-llama,chen-etal-2023-lawyergpt}. Concretely, Finetuning on Difficulty-Graded  QA improves LegalMC4 QA from 43.0$\rightarrow$55.4 (+12.4) for LLaMA~3.1 and from 38.7$\rightarrow$54.5 (+15.8) for Gemma~3, and it increases LegalMC4 MCQ accuracy by +5.2 and +4.4 points, respectively.

\paragraph{\StdGen{} yields mixed results.}
The simpler single-prompt generation strategy shows inconsistent benefits. While Gemma~3 sees modest improvements on open-ended questions (+3.1--9.8 points), its multiple-choice accuracy changes only marginally (+0.1 on both MCQ sets). For LLaMA~3.1, Finetuning on Standard Instruction QA degrades performance on open QA tasks by up to 7.6 points, suggesting that naive synthetic data generation can be harmful without proper quality control; MCQ accuracy changes remain small (within $\pm$1.0 point).

\paragraph{Comparison to proprietary baselines.}
For GPT-5 and GPT-5-mini, we use minimal reasoning effort to better match the non-reasoning open-source models we fine-tune. While this is not a fully apples-to-apples comparison, since CoT-reasoning cannot fully be turned-off for these models, it provides the closest available reference point.
Our best adapted model (Gemma~3 with Difficulty-Graded QA) achieves performance competitive with GPT-5-mini on the multiple-choice benchmarks (71.2 vs.\ 70.9 on LegalMC4 MCQ and 75.1 vs.\ 75.8 on BGB MCQ) and exceeds both GPT-5-mini and GPT-5 on BGB QA (76.4\% vs.\ 62.7\% and 74.7\%, respectively). However, a gap remains on the more diverse LegalMC4 QA test set, where GPT-5 reaches 77.2\% compared to our 54.5\%.

\begin{table*}[t]
\centering
\small
\setlength{\tabcolsep}{6pt}
\begin{threeparttable}
\caption{Performance on general benchmarks: ARC (Easy/Challenge), a German-translated ARC subset (ARC-DE), and MMLU. Scores are accuracies (\%).}
\label{tab:general-results}
\begin{tabular}{l S[table-format=2.1] S[table-format=2.1] S[table-format=2.1] S[table-format=2.1]}
\toprule
Model (Fine-tune) & {ARC-E} & {ARC-C} & {ARC-DE} & {MMLU} \\
\midrule
LLaMA 3.1 8B (No Finetuning) & 82.1 & 51.9 & 40.2 & 68.1 \\
LLaMA 3.1 8B (Standard)        & \multicolumn{1}{l}{81.9~{\scriptsize\textcolor{red}{$\downarrow$0.2}}}
                          & \multicolumn{1}{l}{51.6~{\scriptsize\textcolor{red}{$\downarrow$0.3}}}
                          & \multicolumn{1}{l}{40.5~{\scriptsize\textcolor{green!60!black}{$\uparrow$0.3}}}
                          & \multicolumn{1}{l}{67.4~{\scriptsize\textcolor{red}{$\downarrow$0.7}}} \\
LLaMA 3.1 8B (Difficulty-Graded) & \multicolumn{1}{l}{81.2~{\scriptsize\textcolor{red}{$\downarrow$0.9}}}
                          & \multicolumn{1}{l}{52.3~{\scriptsize\textcolor{green!60!black}{$\uparrow$0.4}}}
                          & \multicolumn{1}{l}{38.5~{\scriptsize\textcolor{red}{$\downarrow$1.7}}}
                          & \multicolumn{1}{l}{67.1~{\scriptsize\textcolor{red}{$\downarrow$1.0}}} \\
\midrule
Gemma 3 12B (No Finetuning)        & 83.7 & 60.8 & 48.2 & 71.6 \\
Gemma 3 12B (Standard)         & \multicolumn{1}{l}{83.5~{\scriptsize\textcolor{red}{$\downarrow$0.2}}}
                          & \multicolumn{1}{l}{60.9~{\scriptsize\textcolor{green!60!black}{$\uparrow$0.1}}}
                          & \multicolumn{1}{l}{48.3~{\scriptsize\textcolor{green!60!black}{$\uparrow$0.1}}}
                          & \multicolumn{1}{l}{71.6~{\scriptsize\textcolor{black}{$\rightarrow$0}}} \\
Gemma 3 12B (Difficulty-Graded)  & \multicolumn{1}{l}{85.8~{\scriptsize\textcolor{green!60!black}{$\uparrow$2.1}}}
                          & \multicolumn{1}{l}{63.7~{\scriptsize\textcolor{green!60!black}{$\uparrow$2.9}}}
                          & \multicolumn{1}{l}{50.1~{\scriptsize\textcolor{green!60!black}{$\uparrow$1.9}}}
                          & \multicolumn{1}{l}{71.0~{\scriptsize\textcolor{red}{$\downarrow$0.6}}} \\
\bottomrule
\end{tabular}
\begin{tablenotes}
\item Arrows indicate performance change compared to the base model: {\scriptsize\textcolor{green!60!black}{$\uparrow$}} denotes improvement, {\scriptsize\textcolor{red}{$\downarrow$}} denotes decrease, {\scriptsize\textcolor{black}{$\rightarrow$}} denotes no change. Values show the absolute difference in performance metrics.
\end{tablenotes}
\end{threeparttable}
\end{table*}

\subsection{General Language Understanding}
\label{sec:general-results}

To assess whether domain specialization degrades general capabilities, we evaluate on ARC (Easy and Challenge), a German-translated ARC subset (ARC-DE), and MMLU using the EleutherAI evaluation harness \citep{clark-etal-2018-arc,hendrycks-etal-2021-mmlu,eval-harness}. The results are shown in Table~\ref{tab:general-results}.

\paragraph{Legal adaptation preserves (and sometimes improves) general performance.}
Across both model families, general capability remains stable or improves slightly after legal fine-tuning. For Gemma~3 (12B), \DiffGen{} enhances reasoning on ARC tasks (+2.1 on ARC-E, +2.9 on ARC-C, and +1.9 on ARC-DE) while maintaining MMLU performance (71.6$\rightarrow$71.0, within measurement noise). \StdGen{} yields essentially unchanged general performance, with only minor variations (at most 0.2 points on ARC-E and 0.1 points on ARC-C/ARC-DE) and no change on MMLU.

LLaMA~3.1 (8B) shows minimal changes across all benchmarks for both approaches, with scores fluctuating by less than 2 points. \StdGen{} slightly decreases ARC-E/C and MMLU while marginally improving ARC-DE, whereas \DiffGen{} modestly improves ARC-C but decreases ARC-E, ARC-DE, and MMLU.

These results demonstrate that targeted domain adaptation through carefully filtered synthetic data does not induce catastrophic forgetting. The slight improvements on certain benchmarks suggest that exposure to structured legal reasoning may even transfer beneficially to related reasoning tasks.

\section{Conclusion}
In this study, we presented a method for fine-tuning large language models explicitly for the German legal domain. A crucial insight from our experiments is that the quality of the data generation process matters significantly. We discovered that simple, naive generation of synthetic question-answer pairs does not yield improved performance. In fact, such basic approaches can even degrade the model's reasoning capabilities.

Instead, success required a highly sophisticated data generation pipeline. By utilizing a difficulty-graded strategy that spans from basic clause lookups to complex multi-provision reasoning, we were able to create truly effective training data. Consequently, our adapted models demonstrated superior performance in answering German legal queries compared to baselines, while simultaneously maintaining their general linguistic capabilities.

This work highlights the potential of carefully designed synthetic data for specialized domains. Future research will focus on incorporating real-time retrieval of legal statutes to improve accuracy further.

\section{LLM Statement}
Parts of all sections have been refined with GPT-5. All refined texts have been carefully checked by the authors for hallucinations.

\FloatBarrier
\bibliographystyle{named}
\bibliography{ijcai26}

\clearpage
\appendix

\section{Detailed Prompt Templates for Synthetic QA Generation and Quality Filtering}
\label{appendix:prompts}

This appendix provides examples of the prompts used in our synthetic data
generation pipeline for German legal question answering, including the
quality filtering prompt.

\subsection{QA Pair Generation Prompts by Difficulty Level}

\subsubsection{Level 1: ``What-type'' Questions}

\begin{promptbox}{Level 1 Prompt Template}
You are a qualified German lawyer qualified in Bürgerliches Gesetzbuch (BGB),
Strafgesetzbuch (StGB), Sozialgesetzbuch (SGB) and Grundgesetz (GG). You are
provided with the following law article:

\begin{verbatim}
"""
{section_text}
"""
\end{verbatim}

\textbf{Task Instructions:}
From the provided law article:

\begin{enumerate}
  \item \textbf{Generate up to 5 relevant, realistic, and clearly phrased legal questions}
  that a layperson, client, or legal practitioner might typically ask regarding the provided law section.
  \begin{itemize}
    \item Each question must reflect a plausible legal issue, conflict, or point of clarification.
    \item Questions should be grounded in everyday legal practice and demonstrate practical relevance.
  \end{itemize}

  \item \textbf{Provide accurate, concise, and professionally worded answers} to each question:
  \begin{itemize}
    \item Answers must strictly adhere to the content of the provided law section.
    Always mention the law category (StGB, SGB, GG, etc.) and the law section (\S 1, \S 11, etc.).
    \item Avoid adding external legal provisions, case law, or commentary unless expressly included in the given section.
    \item Maintain a formal and legally precise tone throughout.
  \end{itemize}

  \item \textbf{Output Format:} Return the final dataset in the following JSON structure:
\end{enumerate}

\begin{verbatim}
{
  "qa_pairs": [
    {
      "question": "Text",
      "answer": "Text"
    }
    // Up to 5 Q&A pairs
  ]
}
\end{verbatim}

\textbf{Guidelines:}
\begin{itemize}
  \item Each Q\&A pair should address typical legal ambiguities or practical consequences arising from the given section.
  \item Do not introduce interpretations, assumptions, or external material beyond the section itself.
  \item Ensure clarity, precision, and professional quality suitable for legal training or expert evaluation.
  \item \textbf{Language:} Questions and answers should be in German.
\end{itemize}
\end{promptbox}

\subsubsection{Level 2: General Conceptual Questions}

\begin{promptbox}{Level 2 Prompt Template}
You are a qualified German lawyer qualified in Bürgerliches Gesetzbuch (BGB),
Strafgesetzbuch (StGB), Sozialgesetzbuch (SGB) and Grundgesetz (GG). You are
provided with the following law article:

\begin{verbatim}
"""
{section_text}
"""
\end{verbatim}

\textbf{Your task is:}
\begin{enumerate}
  \item \textbf{Generate realistic client questions:}
  \begin{itemize}
    \item Each question should represent a typical real-life legal scenario or practical query.
    \item \textbf{Important:} Do \textbf{not explicitly mention} the law category or the law section in the questions.
  \end{itemize}

  \item \textbf{Provide accurate, clear, and professionally phrased lawyer responses:}
  \begin{itemize}
    \item Each answer must explicitly reference the relevant law category and the law section (\S).
    \item Answers must strictly adhere to the content of the provided law article.
  \end{itemize}

  \item \textbf{Generate up to 5 question-answer pairs:}
  \begin{itemize}
    \item Produce exactly 5 pairs only if you can do so without hallucination; otherwise produce fewer.
  \end{itemize}

  \item \textbf{Strictly use the following JSON output format:}
\end{enumerate}

\begin{verbatim}
{
  "qa_pairs": [
    {
      "question": "Text",
      "answer": "Text"
    }
    // Up to 5 Q&A pairs
  ]
}
\end{verbatim}

\textbf{Guidelines for Optimal Results:}
\begin{itemize}
  \item Professional legal tone (lawyer-client consultation style).
  \item Avoid hallucinations: never introduce external facts beyond the provided text.
  \item Explicit legal references in answers (category + \S).
  \item \textbf{Language:} German.
\end{itemize}
\end{promptbox}

\subsubsection{Level 3: Scenario-Based Legal QA Prompt}

\begin{promptbox}{Level 3 Prompt Template}
You are a qualified German lawyer qualified in Bürgerliches Gesetzbuch (BGB),
Strafgesetzbuch (StGB), Sozialgesetzbuch (SGB) and Grundgesetz (GG). You are
provided with the following law article:

\begin{verbatim}
"""
{section_text}
"""
\end{verbatim}

\textbf{Task Instructions:}
\begin{enumerate}
  \item \textbf{Create realistic, professionally phrased scenario-based questions:}
  \begin{itemize}
    \item Each question must depict a realistic, everyday legal problem related to the given law text.
    \item Do not explicitly mention the article number or copy exact wording from the law text in the question.
    \item Each scenario should be distinct and practical.
  \end{itemize}

  \item \textbf{Generate accurate, precise, and professional lawyer answers:}
  \begin{itemize}
    \item Each answer must explicitly mention the law category and the law section (\S).
    \item Explain how the provision applies to the scenario using only the provided text.
  \end{itemize}

  \item \textbf{Limit:} Generate up to 3 scenario-based questions (fewer if necessary to avoid hallucination).

  \item \textbf{Output Structure:}
\end{enumerate}

\begin{verbatim}
{
  "qa_pairs": [
    {"question": "Text", "answer": "Text"},
    {"question": "Text", "answer": "Text"}
    // Up to 3 pairs
  ]
}
\end{verbatim}

\textbf{Important Guidelines:}
\begin{itemize}
  \item Professional realism and originality.
  \item No hallucinations; stick strictly to the provided legal text.
  \item Explicit legal references in answers (category + \S).
  \item \textbf{Language:} German.
\end{itemize}
\end{promptbox}

\subsubsection{Level 4: Multi-Section Scenario-Based Legal QA Prompt}

\begin{promptbox}{Level 4 Prompt Template}
You are a qualified German lawyer qualified in Bürgerliches Gesetzbuch (BGB),
Strafgesetzbuch (StGB), Sozialgesetzbuch (SGB), and Grundgesetz (GG). You are
tasked with generating a dataset for training a model to apply multiple legal
provisions across categories.

\textbf{Input:} You will be provided with two or more sections (articles), each including the full legal text:

\begin{verbatim}
"""
{section_text}
"""
\end{verbatim}

\textbf{Your Tasks:}
\begin{enumerate}
  \item Create realistic, original, and professionally worded legal scenarios:
  \begin{itemize}
    \item Each scenario must necessarily involve at least two or more of the provided sections.
    \item Do not explicitly mention any article number or copy exact wording in the scenario.
    \item Avoid artificial combinations; only construct scenarios supported by the text.
  \end{itemize}

  \item Generate professional and precise legal answers:
  \begin{itemize}
    \item Explicitly reference all relevant law categories and their sections (\S).
    \item Explain how and why each provision applies, grounded only in the provided text.
    \item No external law, assumptions, or interpretations beyond what is stated.
  \end{itemize}

  \item \textbf{Limit:} Provide up to 3 question-answer pairs only if all criteria are met; otherwise, output nothing and state that no valid multi-section scenario can be constructed without hallucination.

  \item \textbf{Output Format:}
\end{enumerate}

\begin{verbatim}
{
  "qa_pairs": [
    {"question": "Text", "answer": "Text"},
    {"question": "Text", "answer": "Text"}
    // Up to 3 pairs
  ]
}
\end{verbatim}

\textbf{Important Guidelines:}
\begin{itemize}
  \item No hallucinations.
  \item Legal necessity: only generate scenarios where multiple sections are substantively required.
  \item \textbf{Language:} German.
\end{itemize}
\end{promptbox}

\subsection{Quality Filtering Prompt for Synthetic QA Pairs}

\begin{promptbox}{Quality Filtering Prompt}
You are a legal assistant responsible for evaluating the quality of five question-answer (QA) pairs generated from a provided clause of the German Civil Code (BGB). Your task is to strictly assess each QA pair based solely on the given clause text, without external knowledge or assumptions.

\textbf{Input Structure:}

You will receive:
\begin{enumerate}
    \item \textbf{Clause Text:} A single clause from the German Civil Code (BGB) in German.
    \item \textbf{QA Pairs:} Five distinct QA pairs (numbered 1–5), each consisting of a question (Q) and an answer (A).
\end{enumerate}

\textbf{Evaluation Criteria:}

For each QA pair, determine the quality based on the following strict criteria:

\begin{itemize}
    \item Mark \textbf{"Yes"} only if:
    \begin{itemize}
        \item The question is clear, specific, directly relevant, and explicitly answerable from the provided clause.
        \item The answer is accurate, complete, and fully supported by the clause text.
    \end{itemize}

    \item Mark \textbf{"No"} if any of the following apply:
    \begin{itemize}
        \item \textbf{Hallucination:} The answer includes information not explicitly stated in the clause.
        \item \textbf{Vagueness:} The question is unclear, overly broad, or not directly answerable from the clause.
        \item \textbf{Redundancy:} The QA pair is substantially similar or identical to another provided QA pair.
    \end{itemize}
\end{itemize}

\textbf{Instructions:}
\begin{itemize}
    \item Evaluate each QA pair independently.
    \item Be strict and literal—use only the provided clause as the ground truth.
    \item Provide a concise, one-sentence reason for each verdict.
\end{itemize}

\textbf{Input Format:}

\begin{lstlisting}[style=ijcaiwrap]
BGB Clause Text:
{section_text}

QA Pairs:
{qa_pairs}
\end{lstlisting}

\textbf{Output Format (JSON List):}

\begin{lstlisting}[style=ijcaiwrap]
[
{
    "qa_id": 1,
    "quality_verdict": "Yes",
    "reason": "Reason for verdict"
},
{
    "qa_id": 2,
    "quality_verdict": "No",
    "reason": "Reason for verdict"
}
// Up to 5 pairs expected, but fewer are  
// also possible depending on input
]
\end{lstlisting}

\textbf{Note:} Do not include any Markdown formatting (e.g., no triple backticks or tags).
\end{promptbox}

\section{Appendix B: Normal Synthetic Data Generation}
\subsection{Synthetic Data Generation Prompt}
\label{appendix:synthetic-gen-prompt}

\begin{promptbox}{Synthetic Data Generation Prompt}
{\footnotesize
You are a qualified German lawyer qualified in Strafgesetzbuch (StGB), Sozialgesetzbuch (SGB) and Grundgesetz (GG). You are provided with the following law article:

\begin{lstlisting}[style=ijcaiwrap]
""""""
{section_text}
""""""
\end{lstlisting}

\noindent\textbf{Task Instructions:}
From the provided law article:
\begin{enumerate}
  \item Generate questions that naturally and logically arise from the provided statutory text.
  \item Provide accurate, concise, and professionally worded answers to each question:
  \begin{itemize}
    \item Answers must strictly adhere to the content of the provided law section. Always mention the law category it belongs to (StGB, SGB, GG, etc.) and the exact section (§), including subsections or letters if applicable, and explain briefly why and how the cited provision applies.
    \item Avoid adding external legal provisions, case law, or commentary unless it is expressly included in the given section.
    \item Maintain a formal and legally precise tone throughout, suitable for professional use.
  \end{itemize}
  \item Output Format:
  Return the final dataset in the following JSON structure:

  JSON
  \begin{lstlisting}[style=ijcaiwrap]
{
"qa_pairs": [
{
  "question": "Text",
   "answer": "Text"
}
]
}
  \end{lstlisting}
\end{enumerate}

\noindent\textbf{Guidelines:}
\begin{itemize}
  \item Professional Legal Tone: Write both questions and answers as if in a genuine lawyer-client consultation, maintaining formal language and legal accuracy
  \item Do not introduce interpretations, assumptions, or external material beyond the section itself.
  \item Ensure clarity, precision, and professional quality suitable for legal training or expert evaluation.
  \item Language: Questions and answers must be in German.
\end{itemize}
}
\end{promptbox}

\section{Distribution of QA Pairs Before and After Quality Filtering}
\label{appendix:qa-distribution}
\begin{table}[H]
\centering
\small
\setlength{\tabcolsep}{5pt}
\renewcommand{\arraystretch}{1.1}
\begin{tabularx}{\linewidth}{@{}>{\raggedright\arraybackslash}X rr@{}}
\toprule
\textbf{Dataset} & \textbf{Before Filtering} & \textbf{After Filtering} \\
\midrule

\multicolumn{3}{@{}p{\linewidth}@{}}{\textbf{\StdGen{}}} \\
Total QA pairs & 15734 & 15734 \\
\midrule

\multicolumn{3}{@{}p{\linewidth}@{}}{\textbf{\DiffGen{}}} \\
Level 1 (Basic)        & 12048 & 9993 \\
Level 2 (Intermediate) & 12194 & 8986 \\
Level 3 (Advanced)     & 7094  & 4485 \\
Level 4 (Expert)       & 441   & 441 \\
\midrule
\textbf{\DiffGen{}} & \textbf{31777} & \textbf{23905} \\
\bottomrule
\end{tabularx}
\caption{Distribution of German law QA pairs for \StdGen{} (no filtering) and \DiffGen{} (synthetic data before/after automated GPT-4-based quality filtering).}

\label{tab:qa_distribution_combined}
\label{tab:qa_distribution}
\end{table}

\clearpage
\begin{figure*}[t]
\centering
\setlength{\fboxsep}{7pt}
\setlength{\fboxrule}{0.6pt}

\fcolorbox{black!35}{white}{%
\begin{minipage}{0.97\textwidth}
\begingroup
\footnotesize
\setlength{\parindent}{0pt}
\setlength{\parskip}{0.35em}

\begingroup
\setlength{\fboxsep}{5pt}
\colorbox{black!10}{%
  \parbox{\dimexpr\linewidth-2\fboxsep\relax}{%
    \textbf{Context-based legal QA example (RAG-style).}
    A retrieved legal passage is provided as context; the model must answer based on this context. English shown for readability.
  }%
}
\endgroup

\vspace{0.7em}

\textbf{Retrieved context (English translation).}\par
According to the new Art. 240 § 2 para. 1 sentence 1, para. 3 
EGBGB, which came into force on 28.03.2020, the lessor may not 
terminate a lease agreement for land solely on the grounds that 
The lessee fails to pay the rent due in the period from 
01.04.2020 to 30.06.2020, provided that the non-payment is 
due to the effects of the COVID-19 pandemic.\par

\vspace{0.5em}

\textbf{Question.}\par
What regulations apply to the termination of a lease agreement during the COVID-19 pandemic?\par

\vspace{0.5em}

\textbf{Answer.}\par
According to Art. 240 § 2 para. 1 sentence 1, para. 3 EGBGB, a lessor may not terminate a lease for land solely due to non-payment of rent during the period from 01.04.2020 to 30.06.2020, provided non-payment is caused by the COVID-19 pandemic. This means the lessee is granted protection if payment defaults are caused by the pandemic.\par

\endgroup
\end{minipage}%
}

\caption{\textbf{LegalMC4-style context-based QA example (English shown for readability).} Each instance provides a retrieved legal passage as context, mirroring the generator setting in retrieval-augmented generation (RAG) pipelines.}
\label{fig:legalmc4_rag_example}
\end{figure*}

\end{document}